%% file: root.tex
\documentclass[10pt, a4paper,conference]{IEEEtran}
\ifCLASSINFOpdf
\else
\fi
\usepackage[table]{xcolor} 
\usepackage{graphicx}
\usepackage{caption}
\usepackage{booktabs}
\usepackage{hyperref}
\usepackage{lipsum}
\usepackage{subfigure}

\usepackage{soul}

\input{tables.tex}

\newcommand{\NetName}{AerialMPTNet}

\begin{document}

\title{\NetName: Multi-Pedestrian Tracking in Aerial Imagery Using Temporal and Graphical Features
}
%SkyTrackerNet: Improving Pedestrian Tracking in Aerial Images by using Temporal and Neighboral Features

%\author{\IEEEauthorblockN{Maximilian Kraus}
%\IEEEauthorblockA{Department of Informatics\\
%Technical University of Munich\\
%Munich, Germany\\
%Email: maximilian.kraus@tum.de}
%\and
%\IEEEauthorblockN{Reza Bahmanyar}
%\IEEEauthorblockA{Remote Sensing Technology Institute\\
%German Aerospace Center (DLR)\\
%Wessling, Germany\\
%Email: reza.bahmanyar@dlr.de}
%\and
%\IEEEauthorblockN{Majid Azimi}
%\IEEEauthorblockA{Remote Sensing Technology Institute\\
%German Aerospace Center (DLR)\\
%Wessling, Germany\\
%Email: seyedmajid.azimi@dlr.de}
%\and
%\IEEEauthorblockN{Peter Reinartz}
%\IEEEauthorblockA{Remote Sensing Technology Institute\\
%German Aerospace Center (DLR)\\
%Wessling, Germany\\
%Email: peter.reinartz@dlr.de}}


\author{\IEEEauthorblockN{Maximilian~Kraus\IEEEauthorrefmark{1}\IEEEauthorrefmark{2},
Seyed~Majid~Azimi\IEEEauthorrefmark{1}\IEEEauthorrefmark{3},
Emec~Ercelik\IEEEauthorrefmark{2},
Reza~Bahmanyar\IEEEauthorrefmark{1},
Peter~Reinartz\IEEEauthorrefmark{1}, and
Alois Knoll\IEEEauthorrefmark{2}}
\IEEEauthorblockA{\IEEEauthorrefmark{1}Remote Sensing Technology Institute, German Aerospace Center (DLR),
Wessling, Germany\\
Emails: \{maximilian.kraus;~seyedmajid.azimi;~reza.bahmanyar;~peter.reinartz\}@dlr.de}
\IEEEauthorblockA{\IEEEauthorrefmark{2}Department of Informatics, \IEEEauthorrefmark{3}Department of Aerospace, Aeronautics and Geodesy,\\Technical University of Munich, Munich, Germany\\
Emails: \{maximilian.kraus;~seyedmajid.azimi;~emec.ercelik;~alois.knoll\}@tum.de}}
%Telephone: (800) 555--1212, Fax: (888) 555--1212}


\makeatletter

\let\@oldmaketitle\@maketitle% Store \@maketitle

\renewcommand{\@maketitle}{\@oldmaketitle% Update \@maketitle to insert...

%\centering\href{}{\color{pink}{}}\\

%  \vspace{-0.6cm}
%\todo{Shouldn't we reference this Figure }
\includegraphics[width=\linewidth,height=8\baselineskip]{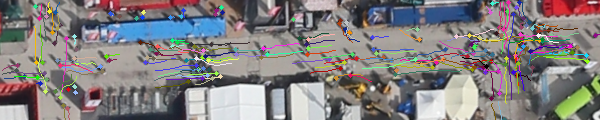}\\% ... an image
\centering
    Sample aerial image with its overlaid annotations from the AerialMPT dataset taken over the BAUMA 2016 trade fair.\bigskip}

\makeatother

\maketitle

\begin{abstract}

Multi-pedestrian tracking in aerial imagery has several applications such as large-scale event monitoring, disaster management, search-and-rescue missions, and as input into predictive crowd dynamic models. Due to the challenges such as the large number and the tiny size of the pedestrians (e.g., $4\times4$ pixels) with their similar appearances as well as different scales and atmospheric conditions of the images with their extremely low frame rates (e.g., 2~fps), current state-of-the-art algorithms including the deep learning-based ones are unable to perform well.
In this paper, we propose \NetName, a novel approach for multi-pedestrian tracking in geo-referenced aerial imagery by fusing appearance features from a Siamese Neural Network, movement predictions from a Long Short-Term Memory, and pedestrian interconnections from a GraphCNN.
In addition, to address  the lack of diverse aerial pedestrian tracking datasets, we introduce the Aerial Multi-Pedestrian Tracking (AerialMPT) dataset consisting of 307 frames and 44,740 pedestrians annotated. We believe that AerialMPT is the largest and most diverse dataset to this date and will be released publicly. 
We evaluate \NetName{} on AerialMPT and KIT~AIS, and benchmark with several state-of-the-art tracking methods. Results indicate that \NetName{} significantly outperforms other methods on accuracy and time-efficiency.

\end{abstract}


% \fi
\IEEEpeerreviewmaketitle

\input{sections/introduction}
\input{sections/related_work}
\input{sections/datasets}

\input{sections/method}

\input{sections/evaluation}

\input{sections/conclusion}
\input{sections/acknowledgment}

\bibliographystyle{IEEEtran}

\bibliography{bibliography.bib}

\end{document}

%% file: tables.tex
\newcommand{\AerialMOTPedestrian}{%
\begin{table}
\vspace*{2mm}

\centering
\caption{Statistics of the train and test sets of the AerialMPT dataset. The image sequences are from different flight campaigns over \textbf{BAUMA} construction trade fair (Munich, Germany), \textbf{OAC} Open Air Concert (Germany), \textbf{Witt} Church day (Wittenberg, Germany), as well as \textbf{Pasing}, \textbf{Marienplatz}, and \textbf{Karlsplatz} Munich city areas (Germany).
}

\resizebox{0.489\textwidth}{!}{%
\begin{tabular}{c||c|c|c|c|c|c}
\toprule
\multicolumn{7}{c}{\textbf{Training}} \\
\rowcolor{gray!50}
Seq. & Image Size & \# Frames & \# Pedestrian & \# Anno. & \# Anno./fr. & GSD (cm)\\
\hline
Bauma1 & 462x306 & 19 & 270 & 4,448 & 234.11 & 11.5  \\ 
Bauma2 & 310x249 & 29 & 148 & 3,627 & 125.07 & 11.5 \\
Bauma4 & 281x243 & 22 & 127  & 2,399 & 109.05 & 11.5 \\
Bauma5 & 281x243 & 17 & 94 & 1,410 & 82.94 & 11.5 \\
Marienplatz & 316x355 & 30 & 215 & 5,158 & 171.93 & 10.5  \\
Pasing1L & 614x366 & 28 & 100 & 2,327 & 83.11 & 10.5 \\
Pasing1R & 667x220 & 16 & 86 & 1,196 & 74.75 & 10.5  \\
OAC & 186x163 & 18 & 92 & 1,287 & 71.50 & 8.0 \\

\hline
\multicolumn{2}{c|}{\textbf{Total}} & 179 & 1,132  & 21,852  & 122.08  \\
\bottomrule
\end{tabular} 
}

\resizebox{0.489\textwidth}{!}{%
\begin{tabular}{c||c|c|c|c|c|c}
\toprule
\multicolumn{7}{c}{\textbf{Testing}} \\
\rowcolor{gray!50}

Seq. & Image Size & \# Frames & \# Pedestrian & \# Anno. & \# Anno./fr. & GSD (cm)\\
\hline
Bauma3 & 611x552 & 16 & 609 & 8,788 & 549.25 & 11.5\\
Bauma6 & 310x249 & 26 & 270 & 5,314 & 204.38 & 11.5\\
Karlsplatz & 283x275 & 27 & 146 & 3,374 & 124.96 & 10.0\\
Pasing7 & 667x220 & 24 & 103 & 2,064 & 86.00 & 10.5 \\
Pasing8 & 614x366 & 27 & 83 & 1,932 & 71.56 & 10.5\\
Witt & 353x1202 & 8 & 185 & 1,416 & 177.00 & 13.0\\
\hline
\multicolumn{2}{c|}{\textbf{Total}} & 128 & 1,396  & 22,888  & 178.81 \\
\bottomrule
\end{tabular} 
}
\label{tab:dataset}

\end{table}
}

\newcommand{\AerialMOTvsKIT}{%
\begin{table*}
\vspace*{2mm}

	\centering
		 \caption{Comparing the statistics of the AerialMPT and KIT~AIS pedestrian tracking datasets.}
		\resizebox{\textwidth}{!}{
		\begin{tabular}{c|c|c|c c| c c c c| c c c | c | c}
		\rowcolor{gray!50}
		\multicolumn{1}{c}{Dataset} & 
		\multicolumn{1}{c}{Avg. } & 
		\multicolumn{1}{c}{GSD} &
		\multicolumn{2}{c}{\# Sequences} &
		\multicolumn{4}{c}{\# Frames} &
		\multicolumn{3}{c}{\# Annotations (per Frame)}&
		\multicolumn{1}{c}{Total} &
		\multicolumn{1}{c}{FPS}\\
		\rowcolor{gray!50}
		 & Image Size & (cm) & {Train} & Test & Total & Avg. & Min & Max & Avg. & Min & Max & Annotations & \\
	    \hline
        KIT~AIS    & 438$\times$483              & 12--15 & 7 & 6  & 189 & 14.1    & 4 & 31  & 192.7     & 8    & 877 & 32,760 & 1--2\\
        AerialMPT~(Ours)  & 425$\times$358 & 8--13 & 8 & 6 & 307 & 21.9 & 8 & 30 & 145.7 & 68 & 556 & 44,740 & 2\\ 
	\end{tabular}
	}
	\label{tab:datasetComp} 

\end{table*}
}

\newcommand{\TotalKITResults}{%
\begin{table*}
\centering
\caption{Results of different tracking methods on the KIT~AIS and AerialMPT datasets.}
\resizebox{\textwidth}{!}{%
\rowcolors{2}{gray!25}{white}
\begin{tabular}{|c||c|ccc|ccc|cccc|cccc|ccc|}
\hline
\rowcolor{gray!50}
Tracker  & Dataset &IDF1$\uparrow$ & IDP$\uparrow$ & IDR$\uparrow$ & Rcll$\uparrow$ & Prcn$\uparrow$ & FAR$\downarrow$ & GT & MT\%$\uparrow$ & PT\%$\uparrow$ & ML\%$\downarrow$ & FP$\downarrow$ & FN$\downarrow$ & ID$\downarrow$ & FM$\downarrow$ & MOTA$\uparrow$ & MOTP$\uparrow$ & MOTAL$\uparrow$ \\

\hline
KCF  & KIT~AIS & 9.0 & 8.8 & 9.3 & 10.3 & 9.8 & 165.56 & 1043 & 1.1 & 53.8 & 45.1 & 11426 & 10782 & 32 & \textbf{116} & -84.9 & \textbf{87.2} & -84.7  \\ 
Medianflow  & KIT~AIS & 18.5 & 18.3 & 18.8 & 19.5 & 19.0 & 144.72 & 1043 & 7.7 & \textbf{55.8} & 36.5 & 9986 & 9678 & \textbf{30} & 161 & -63.8 & 77.7 & -63.5 \\ 
CSRT  & KIT~AIS & 16.0 & 16.9 & 15.2 & 17.5 & 19.4 & 126.55 & 1043 & 9.6 & 51.0 & 39.4 & 8732 & 9924 & 91 & 254 & -55.9 & 78.4 & -55.1 \\ 
Mosse  & KIT~AIS & 9.1 & 8.9 & 9.3 & 10.5 & 10.0 & 163.81 & 1043 & 0.8 & 54.0 & 45.2 & 11303 & 10765 & 31 & 133 & -85.8 & 86.7 & -83.5 \\

Tracktor++  & KIT~AIS & 6.6 & 9.0 & 5.2 & 10.8 & 18.7 & \textbf{81.86} & 1043 & 1.1 & 28.4 & 70.5 & \textbf{5648} & 10723 & 648 & 367 & -41.5 & 40.5 & --   \\ 

Stacked DCFNet  & KIT~AIS & 30.0 & 30.2 & 30.9 & 33.1 & 32.3 & 120.52 & 1043 & 13.8 & 62.6 & 23.6 & 8316 & 8051 & 139 & 651 & -37.3 & 71.6 & -36.1  \\ 
SMSOT-CNN  & KIT~AIS & 32.5 & 31.7 & 33.4 & 35.7 & 33.9 & 121.32 & 1043 & 22.2 & 56.0 & 21.8 & 8371 & 7730 & 135 & 585 & -35.0 & 70.0 & -33.9  \\ 
\NetName{} (LSTM only)  (Ours)  & KIT~AIS & 39.7 & 38.8 & 40.6 & 44.6 & 42.6 & 104.78 & 1043 & \textbf{28.9} & 53.8 & 17.3 & 7230 & 6661 & 270 & 886 & -17.8 & 68.8 & -15.5  \\ 
\NetName{} (GCNN only) (Ours)  & KIT~AIS & 37.5 & 36.7 & 38.4 & 42.0 & 40.0 & 109.49 & 1043 & 25.3 & 55.3 & 19.4 & 7555 & 6980 & 259 & 814 & -23.0 & 69.6 & -20.9   \\ 
\NetName{} (Ours)  & KIT~AIS & \textbf{40.6} & \textbf{39.7} & \textbf{41.5} & \textbf{45.1} & \textbf{43.2} & 103.45 & 1043 & 28.1 & 55.3 & \textbf{16.6} & 7138 & \textbf{6597} & 236 & 897 & \textbf{-16.2} & 69.6 & -\textbf{14.2} \\

\hline
KCF  & AerialMPT & 11.9 & 11.5 & 12.3 & 13.4 & 12.5 & 167.24 & 1396 & 3.7 & 17.0 & 79.3 & 21407 & 19820 & 86  & 212 & -80.5 & 77.2 & -80.1   \\ 
Medianflow  & AerialMPT & 12.2 & 12.0 & 12.4 & 13.1 & 12.7 & 161.97 & 1396 & 1.7 & 20.2 & 78.1 & 20732 & 19883 & \textbf{46} & \textbf{144} & -77.7 & 77.8 & -77.5   \\ 
CSRT  & AerialMPT & 16.9 & 16.6 & 17.1 & 20.3 & 19.7 & 148.52 & 1396 & 2.9 & 37.8 & 59.3 & 19011 & 18235 & 426  & 668 & -64.6 & 74.6 & -62.7   \\ 
Mosse  & AerialMPT & 12.1 & 11.7 & 12.4 & 13.7 & 12.9 & 165.66 & 1396 & 3.8 & 17.9 & 78.3 & 21204 & 19749 & 85  & 194 & -79.3 & \textbf{80.0} & -78.9  \\ 

Tracktor++   & AerialMPT & 4.0 & 8.8 & 3.1 & 5.0 & 8.7 & \textbf{93.02} & 1396 & 0.1 & 7.6 & 92.3 & \textbf{11907} & 21752 & 399 & 345 & -48.8 & 40.3 & --   \\ 

Stacked DCFNet  & AerialMPT & 28.0 & 27.6 & 28.5 & 31.4 & 30.4 & 128.30 & 1396 & 9.4 & 44.2 & 46.4  & 16422 & 15712 & 322 & 944 & -41.8 & 72.3 & -40.4  \\

SMSOT-CNN  & AerialMPT & 32.0 & 30.7 & 33.4 & 36.6 & 33.6 & 129.13 & 1396 & 10.7 & 47.7 & 41.6 & 16529 & 14515 & 359 & 1082 & -37.2 & 68.0 & -35.6  \\

\NetName{} (LSTM only) (Ours)  & AerialMPT & 35.7 & 34.5 & 37.0 & 40.5 & 37.7 & 119.40 & 1396 & 12.8 & 49.8 & 37.4 & 15283 & 13627 & 409 & 1376 & -28.1 & 70.1 & -26.3 \\

\NetName{} (GCNN only) (Ours)  & AerialMPT & 37.0 & 35.7 & 38.3 & 42.0 & 39.1 & 117.05 & 1396 & \textbf{15.6} & 46.0 & 38.4 & 14983 & 13279 & 433 & 1229 & -25.4 & 69.7 & -23.5 \\
\NetName{} (Ours)  & AerialMPT & \textbf{37.8} & \textbf{36.5} & \textbf{39.3} & \textbf{43.1} & \textbf{40.0} & 115.48 & 1396 & 15.3 & \textbf{49.9} & \textbf{34.8} & 14782 & \textbf{13022} & 436 & 1269 & \textbf{-23.4} & 69.7 & \textbf{-21.5}  \\ 
\hline

\hline
\end{tabular} }
\label{tab:totalResultsAMPTD}

\end{table*}
}
\newcommand{\CrossResults}{%
\begin{table*}
\vspace*{2mm}
\centering
\caption{The cross-validation results of \NetName{} on the KIT~AIS and AerialMPT datasets.}
\resizebox{\textwidth}{!}{%
\rowcolors{2}{gray!25}{white}
\begin{tabular}{|cc||ccc|ccc|cccc|cccc|ccc|}
\hline
\rowcolor{gray!50}
Train & Test  & IDF1$\uparrow$ & IDP$\uparrow$ & IDR$\uparrow$ & Rcll$\uparrow$ & Prcn$\uparrow$ & FAR$\downarrow$ & GT & MT$\uparrow$ & PT$\uparrow$ & ML$\downarrow$ & FP$\downarrow$ & FN$\downarrow$ & ID$\downarrow$ & FM$\downarrow$ & MOTA$\uparrow$ & MOTP$\uparrow$ & MOTAL$\uparrow$ \\

AerialMPT  & AerialMPT & 37.8 & 36.5 & 39.3 & 43.1 & 40.0 & 115.48 & 1396 & 15.3 &
 49.9 & 34.8 & 14782 & 13022 & 436 & 1269 & -23.4 & 69.7 & -21.5  \\

KIT~AIS  & KIT~AIS & 40.6 & 39.7 & 41.5 & 45.1 & 43.2 & 103.45 & 1043 & 28.1 & 55.3 & 16.6 & 7138 & 6597 & 236 & 897 & -16.2 & 69.6 & -14.2  \\ 

AerialMPT & KIT~AIS & 20.1 & 19.5 & 20.7 & 24.5 & 23.1 & 142.59 & 1043 & 8.8 & 55.7 & 35.5 & 9839 & 9077 & 193 & 515 & -58.9 & 73.7 & -57.3 \\ 

KIT~AIS & AerialMPT & 19.2 & 18.6 & 19.9 & 22.7 & 21.2 & 150.42 & 1396 & 3.9 & 32.2 & 63.9 & 19254 & 17700 & 301 & 967 & -62.8 & 66.4 & -61.5 \\

\hline
\end{tabular} 
}
\label{tab:crossresults}

\end{table*}
}

\newcommand{\resultsONTestSet}{%
\begin{table*}
\centering
\caption{Results of \NetName{} on the test sets of KIT~AIS and AerialMPT datasets.}

\resizebox{\textwidth}{!}{%
\rowcolors{2}{gray!25}{white}
\begin{tabular}{|c||c|ccc|ccc|cccc|cccc|ccc|}
\toprule
\multicolumn{19}{c}{\textbf{KIT AIS}} \\
\rowcolor{gray!50}
\hline

Evaluation & \# Images & IDF1$\uparrow$ & IDP$\uparrow$ & IDR$\uparrow$ & Rcll$\uparrow$ & Prcn$\uparrow$ & FAR$\downarrow$ & GT & MT$\uparrow$ & PT$\uparrow$ & ML$\downarrow$ & FP$\downarrow$ & FN$\downarrow$ & ID$\downarrow$ & FM$\downarrow$ & MOTA$\uparrow$ & MOTP$\uparrow$ & MOTAL$\uparrow$ \\  

AA\_Crossing\_02 & 13 & 46.7 & 45.6 & 46.9 & 49.3 & 48.8 & 45.08 & 94 & 23.4 & 51.1 & 25.5 & 586 & 576 & 12 & 92 & -3.4 & 69.7 & -2.5 \\
AA\_Walking\_02 & 17 & 41.4 & 40.8 & 42.1 & 43.7 & 42.3 & 93.59 & 188 & 17.0 & 51.6 & 31.4 & 1591 & 1504 & 25 & 231 & -16.8 & 68.5 & -15.9 \\
Munich02 & 31 & 31.2 & 30.2 & 32.3 & 37.8 & 35.3 & 136.77 & 230 & 10.4 & 55.7 & 33.9 & 4240 & 3808 & 192 & 498 & -34.5 & 67.6 & -31.4 \\
RaR\_Snack\_Zone\_02 & 4 & 59.0 & 58.8 & 59.2 & 60.9 & 60.5 & 86.00 & 220 & 33.2 & 65.0 & 1.8 & 344 & 3338 & 4 & 34 & 20.7 & 73.4 & 21.1 \\
RaR\_Snack\_Zone\_04 & 4 & 68.5 & 68.3 & 68.6 & 69.8 & 69.5 & 94.25 & 311 & 45.7 & 51.8 & 2.5 & 377 & 371 & 3 & 42 & 38.9 & 74.2 & 39.1 \\
\hline

Total & 69 & 40.6 & 39.7 & 41.5 & 45.1 & 43.2 & 103.45 & 1043 & 28.1 & 55.3 & 16.6 & 7138 & 6597 & 236 & 897 & -16.2 & 69.6 & -14.2 \\

\hline
\end{tabular} 
}
\vspace{0.2cm}

\resizebox{\textwidth}{!}{%
\rowcolors{2}{gray!25}{white}
\begin{tabular}{|c||c|ccc|ccc|cccc|cccc|ccc|}
\toprule
\multicolumn{19}{c}{\textbf{AerialMPT}} \\
\rowcolor{gray!50}
\hline
Evaluation & \# Images & IDF1$\uparrow$ & IDP$\uparrow$ & IDR$\uparrow$ & Rcll$\uparrow$ & Prcn$\uparrow$ & FAR$\downarrow$ & GT & MT$\uparrow$ & PT$\uparrow$ & ML$\downarrow$ & FP$\downarrow$ & FN$\downarrow$ & ID$\downarrow$ & FM$\downarrow$ & MOTA$\uparrow$ & MOTP$\uparrow$ & MOTAL$\uparrow$ \\  

Bauma3 & 16 & 31.2 & 30.4 & 32.0 & 38.2 & 36.3 & 368.12 & 606 & 11.6 & 51.7 & 36.7 & 5890 & 5435 & 277 & 582 & -32.0 & 70.8 & -28.9 \\
Bauma6 & 26 & 37.2 & 34.8 & 39.9 & 44.2 & 38.6 & 143.69 & 270 & 17.0 & 58.1 & 24.9 & 3736 & 2964 & 123 & 333 & -28.4 & 70.2 & -26.1 \\
Karlsplatz & 27 & 45.6 & 44.2 & 47.1 & 48.6 & 45.6 & 72.37 & 146 & 19.9 & 61.6 & 18.5 & 1954 & 1733 & 25 & 153 & -10.0 & 67.4 & -9.3 \\
Pasing7 & 24 & 67.6 & 64.8 & 70.7 & 71.3 & 65.3 & 32.58 & 103 & 49.5 & 43.7 & 6.8 & 782 & 593 & 5 & 93 & 33.1 & 70.7 & 33.3 \\
Pasing8 & 27 & 39.7 & 38.7 & 40.8 & 41.3 & 39.2 & 45.85 & 83 & 15.7 & 55.4 & 28.9 & 1238 & 1134 & 2 & 83 & -22.9 & 68.9 & -22.8 \\
Witt & 8 & 16.0 & 15.9 & 16.1 & 17.9 & 17.6 & 147.75 & 185 & 2.7 & 24.3 & 73.0 & 1182 & 1163 & 4 & 25 & -65.9 & 60.1 & -65.7 \\
\hline

Total  & 128 & 37.8 & 36.5 & 39.3 & 43.1 & 40.0 & 115.48 & 1396 & 15.3 & 49.9 & 34.8 & 14782 & 13022 & 436 & 1269 & -23.4 & 69.7 & -21.5 \\   
\hline

\end{tabular} 
}
\label{tab:amptresults}

\end{table*}
}

%% file: sections/introduction.tex
\section{Introduction}\label{sec:introduction}

Multi-Object Tracking (MOT) task is to localize multiple moving objects in video frames over time with preserved identity. Despite the progress made in recent years, MOT is still a challenging problem in the computer vision domain due to heavy occlusions and background clutter as well as diverse scales and spatial object densities~\cite{wojke2017simple, bergmann2019tracking, xiang2015learning}.
Despite significant progress on MOT in computer vision using deep learning methods, remote sensing or ``remote vision'' is still in its infancy stage.
MOT on aerial imagery has been challenging to exploit previously, due to the limited level of detail of the images. The development of more advanced camera systems and the availability of very high-resolution aerial images have alleviated the aerial MOT limitations to some extend, allowing a variety of applications ranging from the analysis of ecological systems to aerial surveillance~\cite{remoteSensing2008, everaerts2008use}.
Aerial imagery provides efficient image data over wide areas in a short amount of time. Thus, given sufficient image acquisition speed, developing MOT methods for small moving objects such as pedestrians, vehicles, and ships in image sequences can be investigated to offer new opportunities in disaster management, predictive traffic, and event monitoring.
The large number and the small size of the moving objects together with multiple scales and the very low frame rate (e.g., two fps) are the main differences between MOT in aerial and ground-level datasets.
Besides, the diversity in visibility and weather conditions, as well as the large images and acquisition by moving cameras, add to the complexity of aerial MOT. Despite its important practical application, to the best of our knowledge, only a few research works have dealt with aerial MOT~\cite{reilly2010detection, meng2012object, bahmanyar2019multiple}. 
\begin{figure*}
\vspace*{2mm}
\centering
\includegraphics[width=\textwidth]{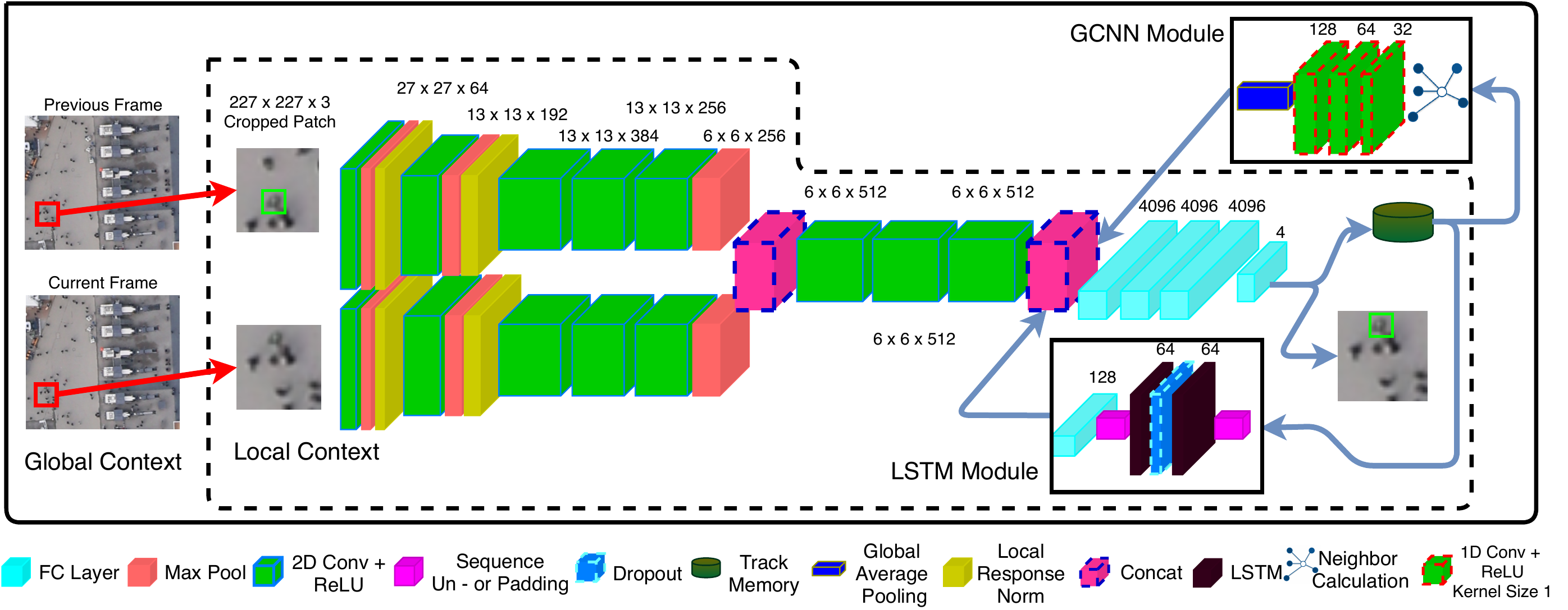}
\caption{Overview of the \NetName's architecture including an SNN, an LSTM, and a GraphCNN module. The inputs are two consecutive image sequences cropped and centered to a target object, and the output is the object coordinates in the second snippet which is mapped to the image coordinates.}
\label{fig:model_overview}
\end{figure*}

Traditional MOT approaches incorporate various methods such as discriminative correlation filters~(DCF)~\cite{henriques2014high}, Kalman and particle filters~\cite{cuevas2005kalman,cuevas2007particle}, and point tracking~\cite{ighrayene2016making}. It has been shown that these methods perform poorly in unconstrained environments due to rigid target modeling and handcrafted target representations~\cite{marvasti2019deep}.
Recently, the rise of Deep Neural Networks (DNNs) has led to significant performance gains in object detection, segmentation, and classification tasks~\cite{he2016deep, szegedy2016rethinking, ren2015faster}. This success also affected visual object tracking, making it possible to develop more robust trackers based on DNNs such as Convolutional Neural Networks (CNNs) \cite{wojke2017simple}, Siamese Neural Networks (SNNs) \cite{held2016learning}, Recurrent Neural Networks~(RNNs) \cite{milan2017online}, and Generative Adversarial Networks (GANs)~\cite{song2018vital}.

In this work, we propose AerialMPTNet as an efficient MOT framework for multi-pedestrian tracking (MPT) in geo-referenced aerial image sequences. AerialMPTNet is a regression-based DNN over the state-of-the-art baseline method of SMSOT-CNN~\cite{bahmanyar2019multiple}. 

\NetName{} is designed and trained so that it incorporates temporal and graphical features of the pedestrian movements for robust and long-term tracking of the pedestrians in various crowd densities and movements. Figure~\ref{fig:model_overview} illustrates an overview of \NetName. Our approach benefits from a long short-term memory (LSTM)~\cite{hochreiter1997long} and a GraphCNN (GCNN) for movement prediction and modeling the interconnections between pedestrians, respectively.
In contrast to the previous works using an individual LSTM for each object to predict the path of multiple objects ~\cite{alahi2016social, cheng2018pedestrian}, \NetName{} uses only one LSTM module to provide a general path prediction based on the last total predictions of the network itself to reduce training and runtime complexity.

Unlike ground imagery enjoying large and diverse annotated MOT datasets such as MOT17~\cite{milan2016mot16}, remote vision is lacking similar datasets for aerial imagery, which limits the development of MOT methods. To the best of our knowledge, the only existing aerial pedestrian tracking dataset is the KIT~AIS\footnote{https://www.ipf.kit.edu/code.php} dataset, which comprises 189 frames with the frame rates of 1--2 fps and 32,760  pedestrians annotated. The images were provided by the German Aerospace Center~(DLR) from their various flight campaigns. The dataset suffers from low-quality annotation and a low degree of diversity.

\begin{figure*}[t!]
\vspace*{2mm}
\centering
\includegraphics[width=.99\textwidth]{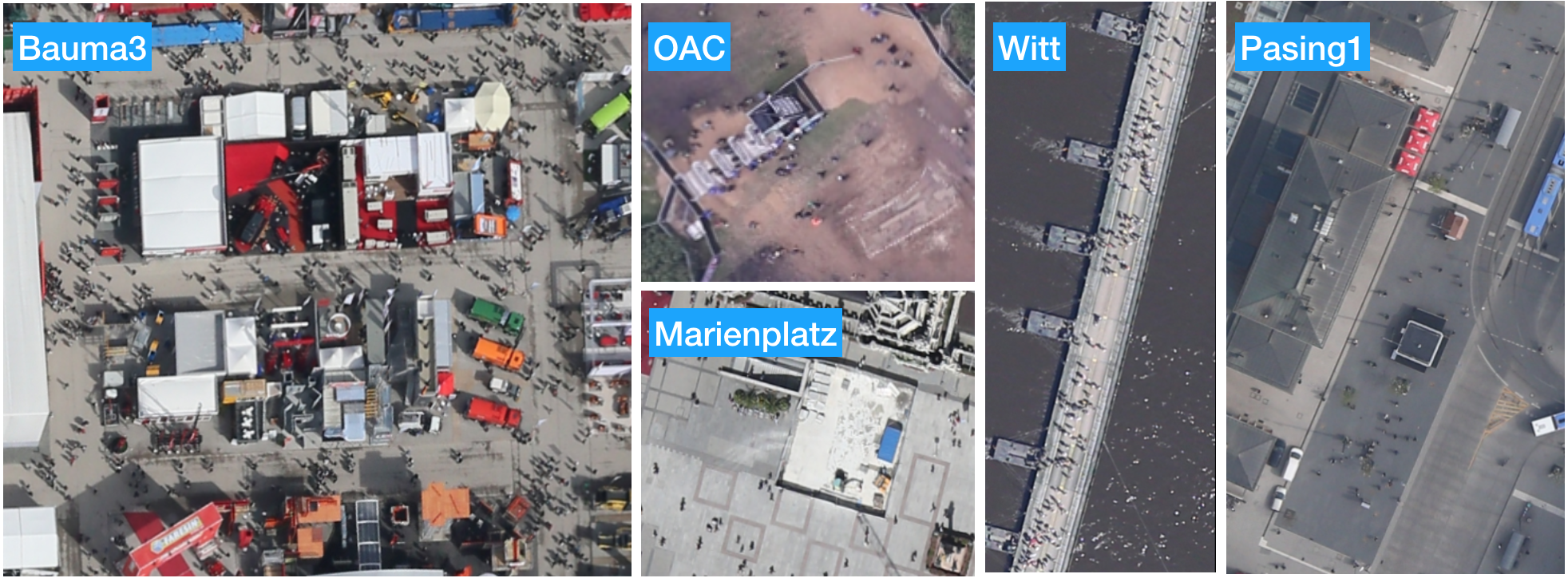}
\caption{Sample images of the AerialMPT dataset from different locations and with various crowd and movement complexities.}
\label{fig:samples}
\end{figure*}

Dealing with the limitations of the KIT~AIS dataset, we introduce the Aerial Multi-Pedestrian Tracking (AerialMPT) dataset, an aerial imagery dataset for pedestrian tracking composed of 307 frames and 44,740 total pedestrians annotated with the frame rate of 2. The DLR's 3K camera system took the image sequences during different flight campaigns,  captured from different crowd scenarios, i.e., from densely moving pedestrian in mass events to the sparse ones in streets. Figure~\ref{fig:samples} demonstrates example images from AerialMPT.
We believe that AerialMPT with its crowd and movement diversity can promote research on aerial MOT. The dataset will be released publicly.

We conduct an intensive qualitative and quantitative evaluation of our approach on the AerialMPT and KIT~AIS datasets. Furthermore, we benchmark the AerialMPT by various traditional and DNN-based methods. Results show that our \NetName{} outperforms all existing methods in the long-term tracking of pedestrians in aerial image sequences. Also, qualitative evaluations demonstrate that \NetName{} overcome the limitations of the existing methods (e.g., ID switches and losing track of objects) by fusing the temporal and graphical features.

In the following, Section 2 provides an overview of existing aerial tracking approaches. In Section 3, we introduce the benchmark datasets.
 Afterwards, we present our methodology in Section 4. In Section 5, we evaluate the proposed method and discuss it. We conclude this paper with Section 6 and give some ideas for future work.

%% file: sections/related_work.tex
\section{Related Work}\label{sec:related_work}
For images taken from airborne platforms, most tracking methods are based on moving object detection \cite{reilly2010detection, benedek2009detection, meng2012object}. For instance, Reilly et al.~\cite{reilly2010detection} eliminated camera motion by a point correspondence based correction method, afterwards motion can be detected by modeling a median background image out of several frames. Images are divided into overlapping cells, and objects are tracked in each cell using the Hungarian algorithm. The cell placement makes it possible to track a large number of objects with \(O(n^3)\). Meng et al.~\cite{meng2012object}  calculate an accumulative difference image from time step to time step to detect moving objects. An object is afterwards modeled by extracting spectral and spatial features. Given a target model, matching candidates can be found in the following frames via regional feature matching. However, such approaches have several disadvantages in our scenario. In general, these methods are sensitive to changing light conditions and the parallax effect, working not well with small or static objects. Reilly et al. use a road orientation estimate as a constraint to assign labels. In our scenario, pedestrians do not walk on predetermined paths such as highways or roads and show more complex moving behaviors. Hence, such estimates can not be used.

Appearance-based methods successfully overcome these issues by working on single images~\cite{liu2015fast, qi2015unsupervised}, especially successful with big objects such as ships, airplanes on the ground, or cars. There is a huge amount of literature covering the topic of pedestrian tracking in surveillance scenarios~\cite{wang2015visual, zhang2016robust}; however, for pedestrian tracking in aerial imagery, the amount of literature is minimal ~\cite{ bahmanyar2019multiple, Schmidt2011}. Schmidt et al.~\cite{Schmidt2011} propose a tracking-by-detection framework based on Haar-like features. Due to different weather conditions and visibilities and the small size of the objects, pedestrians are hardly visible sometimes. Those difficulties result in the regular occurrence of false positives and negatives, influencing the tracking performance negatively. Bahmanyar et al.~\cite{ bahmanyar2019multiple} introduced SMSOT-CNN in 2019, the only previous work dealing with multi-pedestrian tracking in aerial imagery by using deep learning (DL). They extend the single object tracker GOTURN~\cite{held2016learning} with three additional convolution layers and modify the network to be capable of MOT. GOTURN is a regression tracker based on SNNs to track generic objects at high speed. The network receives two image crops as input, one crop from the previous frame centered at the known object position, and one crop from the current frame centered at the same position. A hyperparameter controls the size of the crop, and with this, the amount of context the network obtains. In a final step, the network regresses the object position in crop coordinates. Bahmanyar et al. evaluate SMSOT-CNN on the KIT~AIS pedestrian dataset, reaching a MOTA and MOTP score of -29.9 and 71.0, respectively. However, the network has problems to deal with crowded situations and objects sharing similar appearance features happen to be in the same crop, resulting in identity switches and loosing of tracks.

%% file: sections/datasets.tex
\section{Aerial Multi-Pedestrian Tracking Dataset}\label{sec:datasets}

AerialMPT is an aerial pedestrian tracking dataset composed of 14 sequences and 307 frames of average size $425\times 358$ pixels. The images were taken by the DLR's 3K camera system composed of a nadir-looking and two side-looking commercial DSLR cameras, mounted on a helicopter flying at different altitudes ranging from 600~m to 1400~m. The different flight altitudes resulted in various spatial resolutions (ground sampling distances -- GSDs) from 8 cm/pixel to 13 cm/pixel. 
Due to the movement of the helicopter, the camera system is constantly moving. Therefore, in a post-processing step, for each region of interest, the images were co-registered, geo-referenced, and cropped, resulting in sequences of 2~fps from the region of interest.
The images were acquired at different flight campaigns over various scenes, containing pedestrians, and with different crowd density and movement complexities between 2016 and 2017.
\autoref{fig:samples} demonstrates some sample images from the AerialMPT dataset.

\subsection{Pedestrian Annotation}
The dataset was labeled manually with point-annotations on individual pedestrians by qualified specialist staff, where each individual got assigned a unique ID over the whole sequence. This process resulted in 2,528 pedestrians annotated with 44,740 annotation points, ranging from 71.5 to 549.2 average annotations per frame in the sequences. Since the number of pedestrians in the frames of a sequence could be different (due to entering and leaving pedestrians), we use the annotation effort by the average annotation per frame for each sequence. The annotations were sanity checked by the authors in order to provide precise and accurate annotations.
Pedestrian tracking annotation in aerial imagery is a challenging task due to the large number and the small size of the pedestrians in the images. Due to the similar appearance of the pedestrians, discriminating each person from adjacent pedestrians and similar-looking objects as well as rediscovering the pedestrians occluded for a few frames are difficult and time-consuming.
We split the dataset manually into 8 train and 6 test sequences, where the splits were not randomized. This procedure allowed us to cover all scenes in our train/test splits so that images from the same campaign are either in the training or in the test set.
\autoref{tab:dataset} details the statistics of the image sequences.
\AerialMOTPedestrian

\subsection{Contributions of AerialMPT over KIT~AIS}

\begin{figure}
\centering
\subfigure{\includegraphics[width=.24\textwidth]{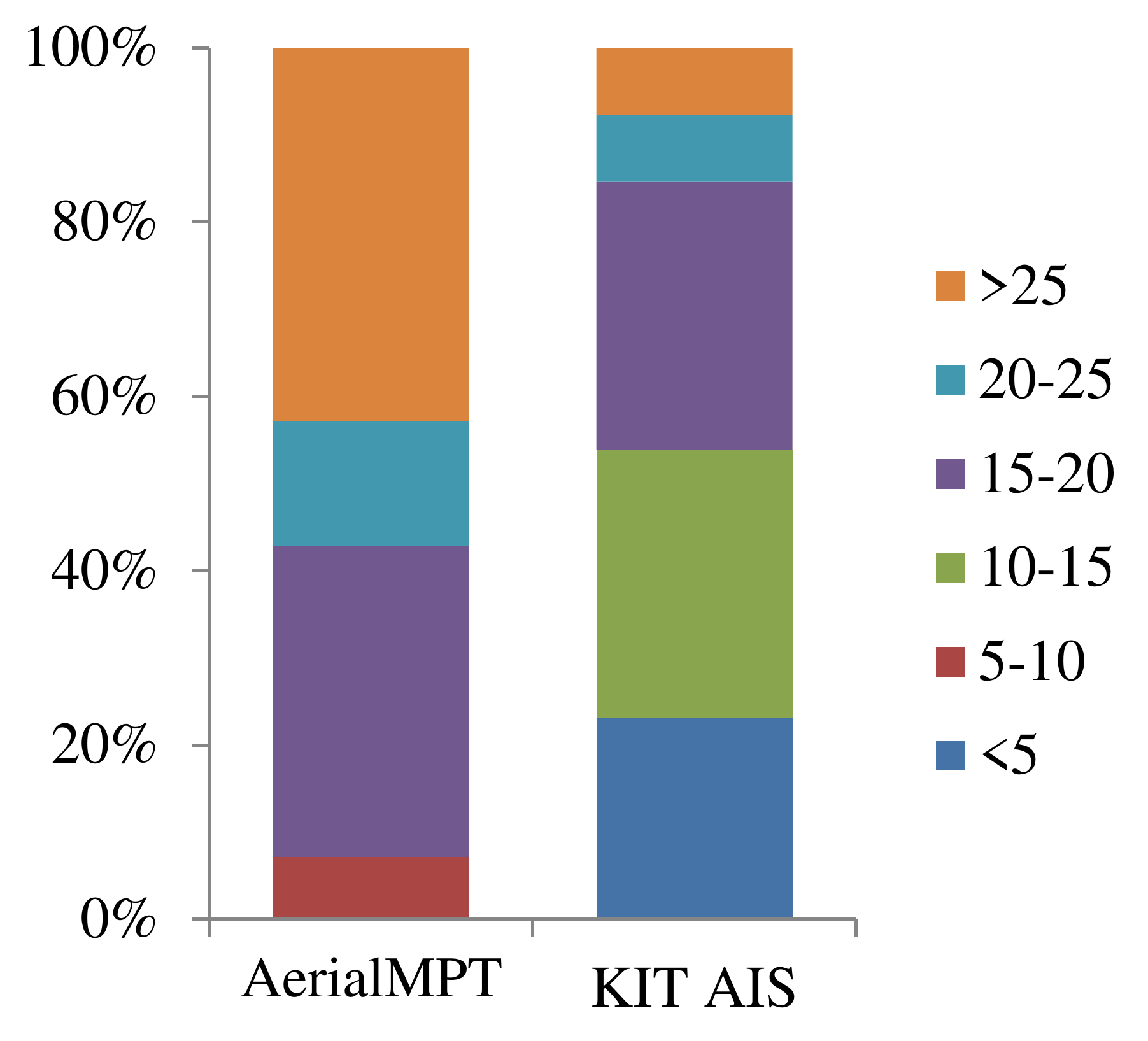}}
\subfigure{\includegraphics[width=.24\textwidth]{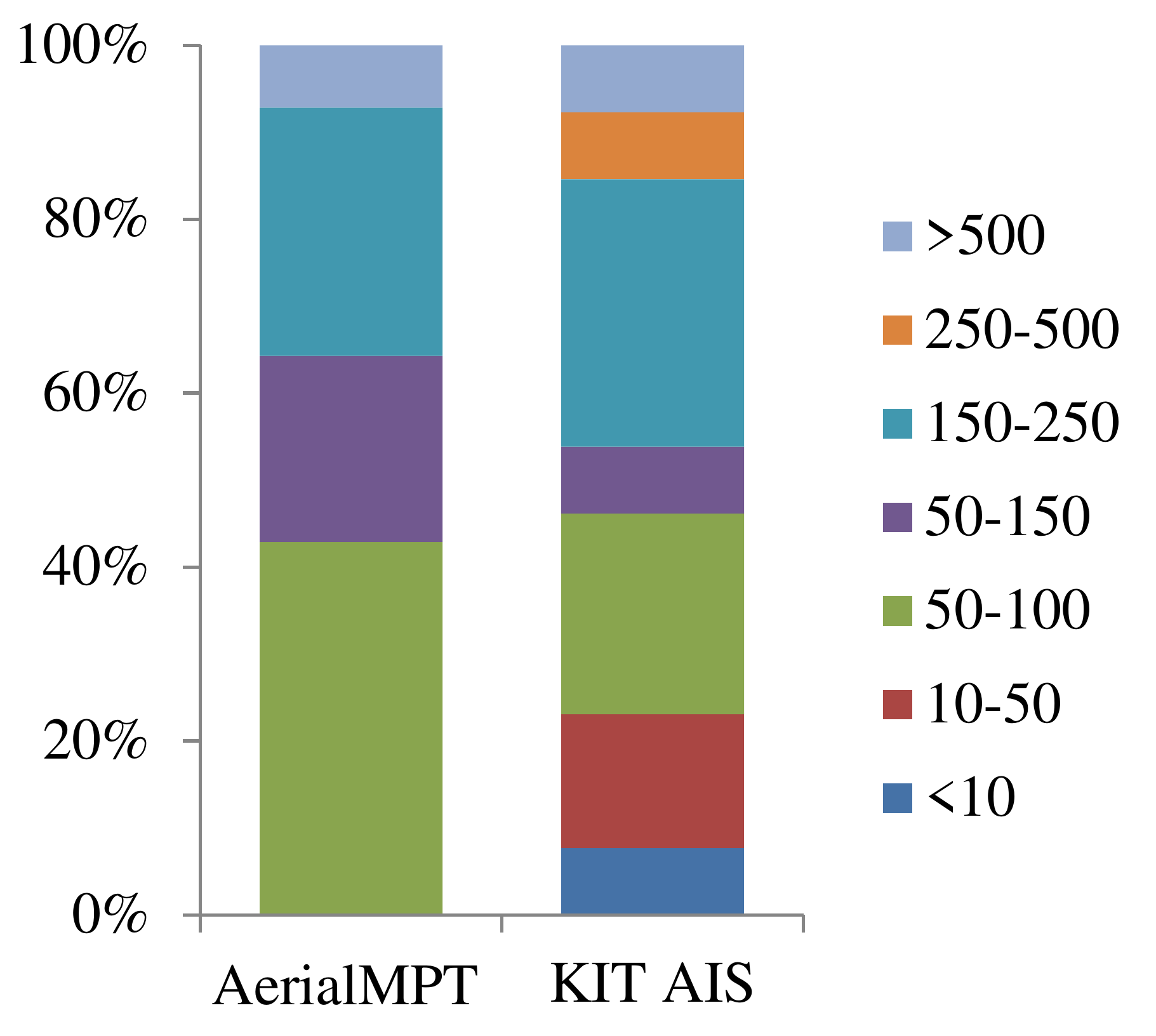}}

\caption{Distribution of (left) number of frames and (right) average per-frame annotations of our AerialMPT and the KIT~AIS pedestrian datasets.}
\label{fig:dataset_statistics}
\end{figure}

The only existing aerial pedestrian tracking dataset is the  KIT~AIS dataset comprising 13 sequences and 189 frames. \autoref{tab:datasetComp} and \autoref{fig:dataset_statistics} compare the statistics of our AerialMPT and the KIT~AIS datasets. As it can be seen, the sequences of AerialMPT usually hold a higher amount of frames than those of KIT~AIS, i.e., 60 \% of the sequences in AerialMPT contain more than 20 frames whereas in KIT~AIS less than 20 \% of the sequences are within this length. The longer sequence length makes AerialMPT more appropriate for long-term pedestrian tracking applications compared to KIT~AIS.
Moreover, the image contrast and quality in AerialMPT is much higher than in KIT~AIS,  which helps tracking methods to discriminate pedestrians and similar-looking objects better.
\AerialMOTvsKIT

Besides, according to \autoref{fig:dataset_statistics}, the scenes in AerialMPT are more crowded and complex than those of KIT~AIS, i.e., all image sequences of AerialMPT contain at least 50 pedestrians; however, more than 20\% of the sequences of KIT~AIS contains less than 10 pedestrians. Furthermore, based on a visual inspection, the spatial densities of the pedestrians in AerialMPT are higher, and their movements are more complex and realistic than those of KIT~AIS. In KIT~AIS the sequences represent simplified and unrealistic movement patterns, e.g., many sequences include only a few tracks ($<$10), or in many scenes, all pedestrians move in the same direction. Altogether, the models trained on AerialMPT can better generalize to various real-world pedestrian tracking scenarios.

%% file: sections/method.tex
\section{Methodology}\label{sec:method}
During our experiments, we noticed that the pedestrians' trajectories are influenced by their previous movements, scene arrangement, and other moving objects. 
However, the current regression-based tracking networks such as SMSOT-CNN do not consider movement history or the relationships between neighboring objects. The networks rely only on a particular neighborhood of the target object (no contextual information outside of the neighborhood). Moreover, inside the neighborhood, the networks do not learn to distinguish the target object from other similar-looking objects. As a consequence, identity switches occur in crowded situations and object crossings. Besides, object tracks could be lost due to background clutter and occlusion.

Dealing with these issues, we propose \NetName{}, which considers the track history and the neighboring objects' interconnections together with the networks' appearance features (see~\autoref{fig:model_overview}).
\NetName{} crops two image tiles from two consecutive frames, namely target and search area, in which the object location is known and to be detected, respectively.
Both tiles are cropped from the same coordinates, centered on the target object, and scaled to 227$\times$227 pixels. Afterwards, they are given to the SNN module, which is composed of two branches of five 2D convolution, three max-pooling, and two local response normalization layers each, where the layer weights are shared between the branches. \autoref{fig:model_overview} details the layer information. 
The output features of the branches $Out_{SNN}$ are then concatenated and given to four fully connected layers that regress the object coordinates in the search area (top-left and bottom-right points of the bounding box around the object).
The predicted object coordinates are then input to the LSTM and GCNN modules.

\subsection{Long Short-Term Memory Module}

The LSTM module is composed of two LSTM layers. For each object being tracked, the network prediction is used to generate a sequence of motion vectors. In our experiments, each track has a dynamic history \(l_i\) of up to five last predictions. Since the tracks do not necessarily start at the same time, the length of the histories can differ, which is considered for padding the tracks and making it possible to be processed as a batch. The padded tracks are fed to the first LSTM layer with a hidden size of 64. The hidden state of the first LSTM layer \(h_t^{l-1}\) goes through a dropout layer with \(p=0.5\) and is given as input to the second LSTM layer.
After that, the output features \(h_t\) of the second LSTM layer are given to a linear layer of size 128. 
Finally, the output of the LSTM module \(Out_{LSTM}\) is concatenated with \(Out_{SNN}\) and \(Out_{Graph}\), the output of the GCNN module. The concatenation allows the network to predict object locations based on a fusion of appearance and movement features.

\subsection{GraphCNN Module}

The GCNN module is composed of three convolution layers with $1\times1$ kernels, and the output channel numbers of 32, 64, and 128.
In order to generate the target object's adjacency graph, based on the location estimates of all objects, eight closest neighbors in a neighborhood of 7.5~m of the object are considered and represented as a directed graph by a set of vectors. The vectors are zero-padded if less than eight neighbors are found.
The track length is limited to five, and a padding procedure applied, similar to the LSTM module.
This graph contains the (\(x\),\(y\)) coordinates of the target object in the image tile coordinate system, and the (\(x\),\(y\)) information of the vectors to the eight selected neighbors. Thus, for each track, the input to the GraphCNN is a matrix of $18\times5$, which is given to the network as a batch of multiple tracks.
The output features of the last convolution layer are gone through a global average pooling to generate the final output \(Out_{Graph}\) of 128 dimensions, which is concatenated to \(Out_{SNN}\) and \(Out_{LSTM}\).

%% file: sections/evaluation.tex
\section{Results and Discussion}\label{sec:evaluation}
In this section, we evaluate our \NetName{} on the AerialMPT and KIT~AIS datasets, and compare its results to a set of traditional methods such as KCF~\cite{henriques2014high}, Medianflow~\cite{kalal2010forward}, Mosse~\cite{bolme2010visual} and CSRT~\cite{lukezic2017discriminative}, and DNN-based methods such as Tracktor++~\cite{bergmann2019tracking}, DCFNet~\cite{wang2017dcfnet}, and SMSOT-CNN~\cite{bahmanyar2019multiple}.

\subsection{Experimental Setup}
We used \textit{Titan XP} GPUs and \textit{PyTorch} for all of our experiments. All networks were trained with an SGD optimizer and an initial learning rate of \(10^{-6}\). However, for the training of SMSOT-CNN, we assigned different fractions of the learning rate to each layer, similar to its original implementation in \textit{Caffe} inspired by the GOTURN's~\cite{held2016learning} implementation\footnote{https://github.com/nrupatunga/PY-GOTURN}. Weight and bias initialization was also identical to the \textit{Caffe} version.
For the training of \NetName{} and SMSOT-CNN, firstly, the SNN module and the FC layers were trained on the DLR-ACD~\cite{bahmanyar2019mrcnet} and tracking datasets simultaneously (similar to~\cite{bahmanyar2019multiple}). Then for \NetName{}, using the model weights, all network modules were trained as a whole on the tracking dataset.  
The learning rate was decayed by a factor of \(0.1\) in every 20K iterations.
For all trainings, the $\mathcal{L}1$ loss was used, \(L(x,y) =
|x - y|\), where \(x\) and \(y\) are the output of the network and ground truth, respectively. 

SMSOT-CNN is trained offline in which the network learns to regress the object location based on only one time step. 
\NetName{} is trained in an end-to-end fashion by using feedback loops to integrate historical movement and interconnection information from previous time steps.
In more detail, a batch of 150 tracks was selected, starting at random time steps between \(0\) and the individual track end \(t_{end}-1\). For each track in the batch, the network's position estimates were stored. The position estimates were given to the LSTM module as a sequence of movement vectors with a length of up to 6 previous network estimates. 
The neighbor calculation for the GCNN is also based on the network's predictions. We searched for nearest neighbors based on the network's position estimates and the true positions of all objects in the specific sequence and frame known from the annotations. If the network failed to track a pedestrian and it moved out of the predicted search window, we removed the object from the batch and inserted a new random track.

\subsection{Evaluation Metrics}
We report all of the widely used metrics in the MOT domain \cite{milan2016mot16}. However, we mainly use MOTA and MOTP in our discussion as the commonly-used metrics for MOT performance evaluation. MOTP describes the capability of a tracker in estimating precise object locations:
\begin{equation}
MOTP = \frac{\sum_{t,i} d_{t,i}}{\sum_t c_t},
\end{equation}
where $d_{t,i}$ is the location error for the matched object \(i\) in frame
\(t\), and $c$ is the total number of matched objects. A tracklet and an annotation are associated as matched if their Intersection over Union (IoU) is greater than 0.5.

\begin{figure*}
\vspace*{2mm}

\centering
\includegraphics[width=.99\textwidth]{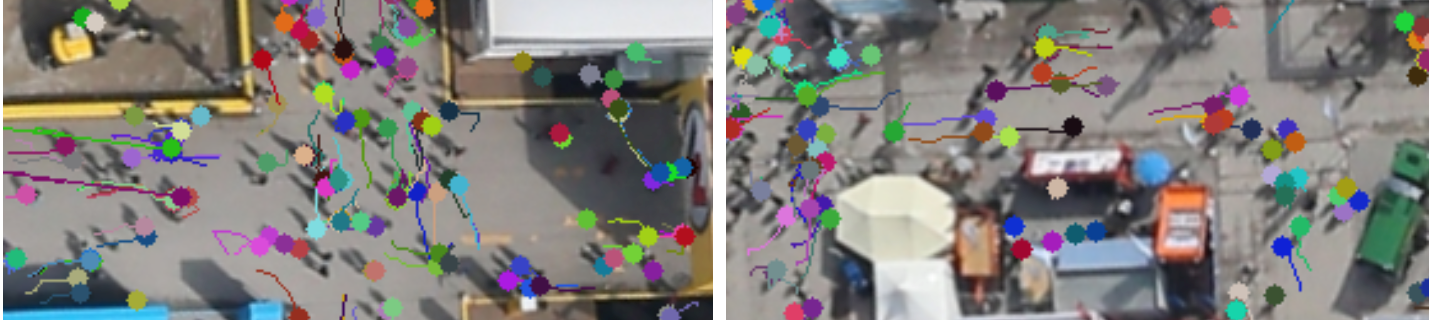}
\caption{Sample predictions of \NetName{} on the Bauma3 (left; 20th frame) and Bauma6 (right, 7th frame) sequences. }
\label{fig:qualResults}
\end{figure*}
\TotalKITResults

MOTA gives an intuitive measure of the tracker's performance at keeping trajectories, independent of the precision of the estimates. It is calculated by summing the false negatives, false positives, and identity switches over all frames divided by the total amount of objects:
\begin{equation}
MOTA = 1 - \frac{\sum_t (FN_t + FP_t + ID_t)}{\sum_t GT_t}.
\end{equation}

In our evaluations, the objects are either Mostly Tracked (MT) – tracked successfully for $>80\%$ of the lifetime, Mostly Lost (ML) – tracked successfully for $<20\%$ of the lifetime, and Partially Tracked (PT) – the rest of the cases. For the other used abbreviations, we refer the readers to~\cite{bahmanyar2019multiple}.

\subsection{Results}

\autoref{fig:qualResults} demonstrates tracking results of our \NetName{} on two sequences of the AerialMPT dataset.
\autoref{tab:totalResultsAMPTD} shows the quantitative results of different tracking methods on the KIT~AIS and the AerialMPT datasets. In general, the DNN-based methods outperform the traditional ones, with MOTA varying between -16.2 and -48.8 versus between -55.9 and -85.8.
Furthermore, CSRT is the best performing traditional methods on both datasets based on MOTA (-55.9 and -64.6). It tracks 9.6\% and 2.9\% of the pedestrians mostly on the KIT~AIS and AerialMPT datasets, while it mostly loses 39.4\% and 59.3\% of the pedestrians, respectively.

According to the table, our \NetName{} outperforms all other methods on both datasets by the significantly highest MOTA (-16.2 and -23.4) and competitive MOTP (69.6 and 69.7) values. It mostly tracks 28.1\% and 15.3\% of the pedestrians (on the two datasets) while mostly loses only 16.6\% and 34.8\% of them.
Among the previous DNN-based methods, SMSOT-CNN achieves the most promising results on both datasets (MOTA: -35.0 and -37.2; MOTP: 70.0 and 68.0). 
DCFNet is a single object tracker originally; however, we adapted its framework to handle multi-object tracking. Although it outperforms SMSOT-CNN in terms of MOTP by 1.6 and 4.3 points, its MOTA values fall behind by 2.4 and 4.6 points.
Tracktor++ is a tracking method based on FasterRCNN~\cite{ren2015faster}. It is the worst-performing among the other DNN-based methods due to suffering from a high amount of FNs and ID switches.

\begin{figure*}
\vspace*{2mm}
\centering
\subfigure{
    \includegraphics[height=2.1cm]{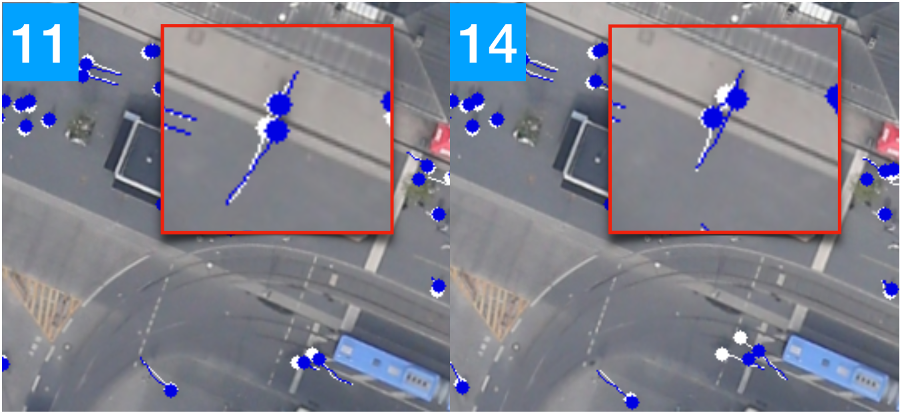}
    \includegraphics[height=2.1cm]{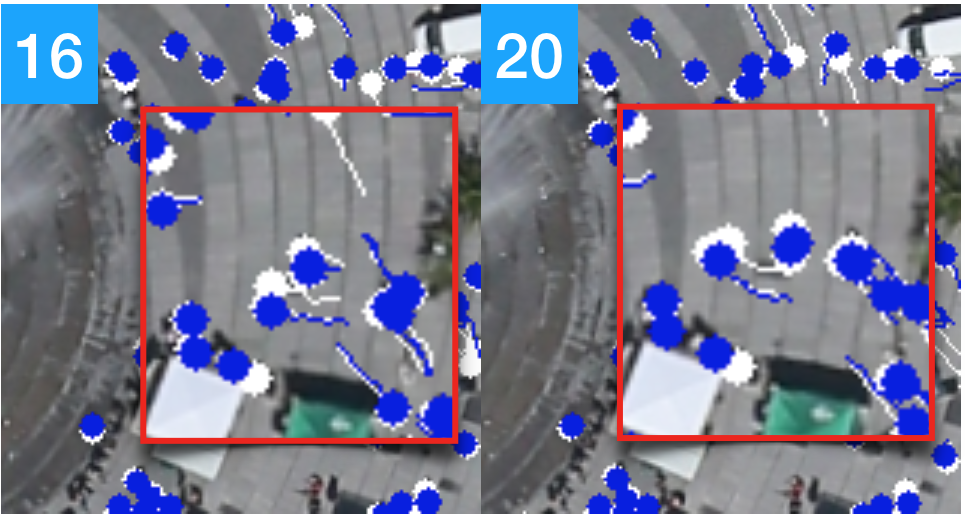}
    \includegraphics[height=2.1cm]{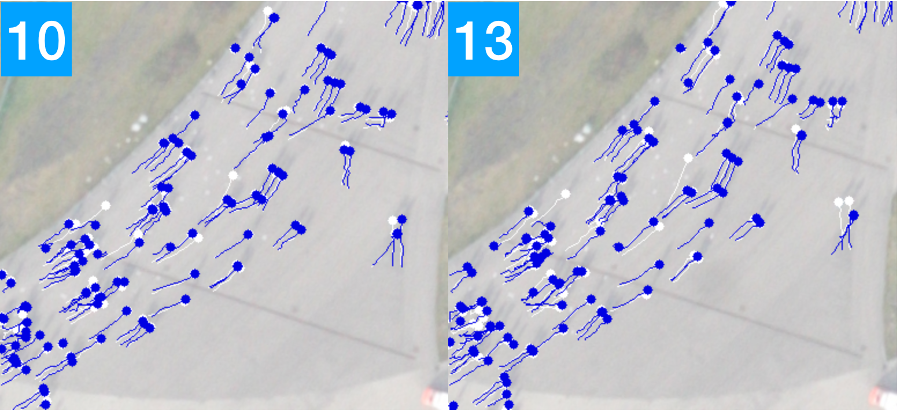}
    \includegraphics[height=2.1cm]{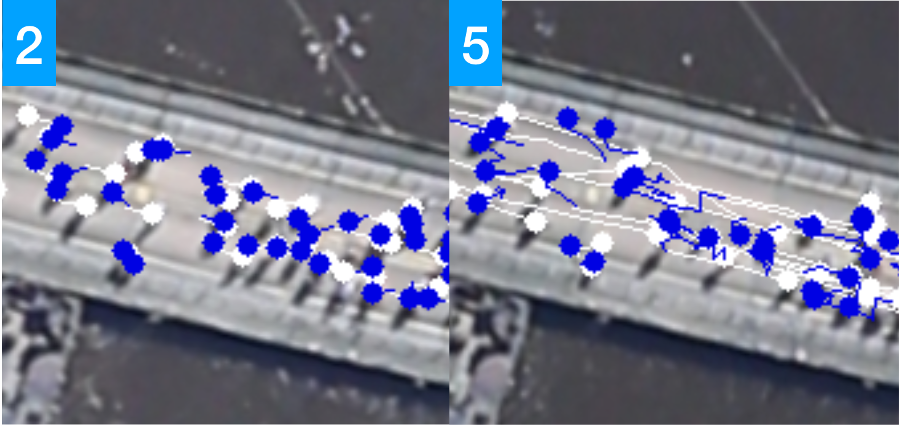}}
\subfigure{
    \includegraphics[height=2.1cm]{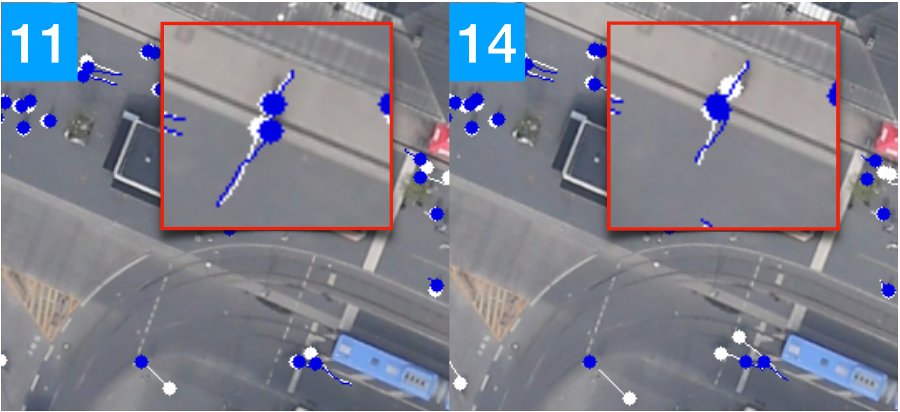}
    \includegraphics[height=2.1cm]{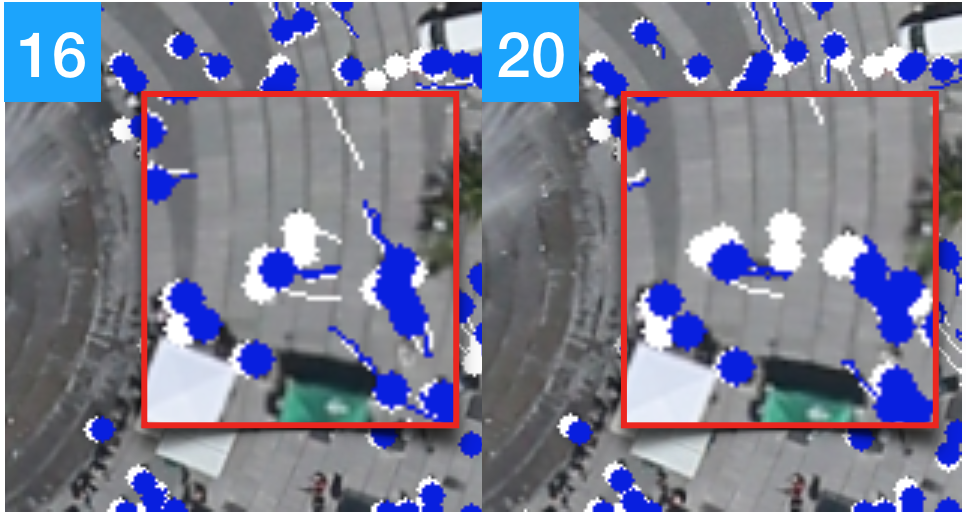}
    \includegraphics[height=2.1cm]{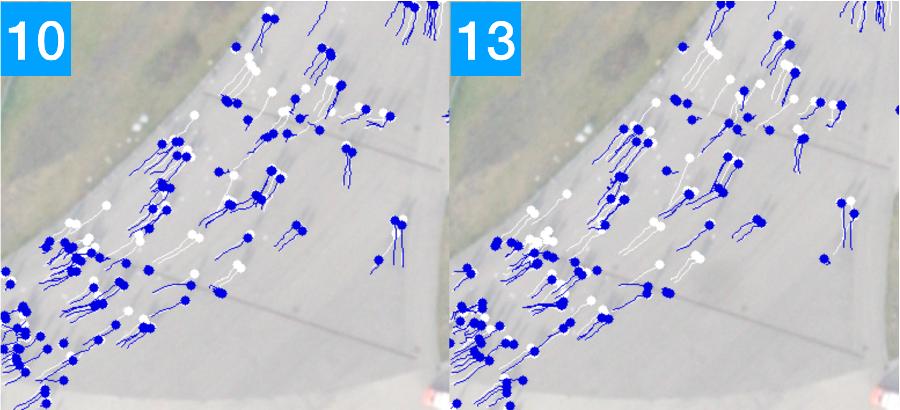}
    \includegraphics[height=2.1cm]{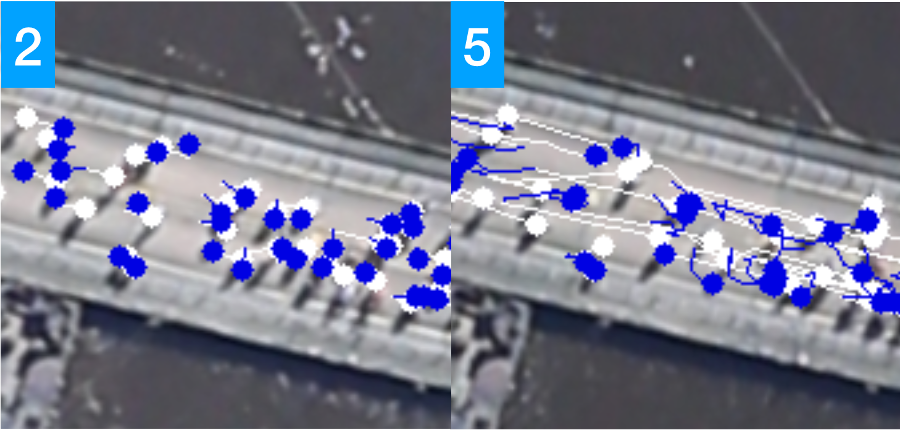}}
\caption{Comparing the performance of \NetName{} (top row) and SMSOT-CNN (bottom row). The first three columns illustrate the outperformance of \NetName{} in pedestrian intersections and keeping the trajectories. The third column shows the poor performance of both methods due to nonadaptive search window size. Samples are from the AerialMPT (Pasing8, Karlsplatz, and Witt sequences) and KIT~AIS (AA\_Walking\_02 sequence) datasets. The frame numbers are also depicted.}
\label{fig:qualComp}
\end{figure*}

Due to the similarity of the SSN module of \NetName{} to SMSOT-CNN, we consider the prediction results of SMSOT-CNN (which only utilizes appearance features) as the baseline for the ablation study of \NetName{}.
According to~\autoref{tab:totalResultsAMPTD}, adding the LSTM module to the SNN module, improves the baseline MOTA by 17.2 and 9.1 points on the KIT~AIS and AerialMPT datasets, respectively.
Moreover, adding the GCNN module to the SNN module improves the baseline MOAT by 12 and 11.8 points. According to the results, considering both modules increases the baseline MOTA by 18.8 and 13.8 points.
\autoref{fig:qualComp} compares the baseline SMSOT-CNN and our \NetName{} on sample sequences of the two datasets qualitatively. According to the results of the first three columns, the fusion of the appearance, temporal, and graphical features allows our \NetName{} to outperform the baseline by better handling the pedestrian crossing situations (avoid ID switches) and keeping the pedestrian trajectories for a longer-term even in the presence of interrupting features.

\autoref{tab:amptresults} shows the tracking results of our \NetName{} on the test sequences of KIT~AIS and AerialMPT datasets. 
According to the table, on the KIT~AIS dataset, the tracker usually achieves better MOTA and MOTP scores for the shorter sequences. On this dataset, the worst MOTA (-34.5) is obtained on the \textit{Munich02} sequence, which is the most complex sequence considering its length as well as the number of pedestrians and their movements. 
On the AerialMPT dataset, the MOTA scores are not correlated to the sequence lengths. This finding indicates that the scene complexities are well-distributed over different sequences of AerialMPT. According to the results, despite its small number of frames, the MOTA score of the \textit{Witt} sequence is relatively low (-65.9). Further investigation shows that the poor performance is caused by the search-window dependency of \NetName{} on the size of the tracked object. In the \textit{Witt} sequence, due to its very different GSD, the objects move out of the search-window (and are therefore lost) at some point, which influences the tracking results negatively.
\resultsONTestSet

In order to demonstrate how an approach trained on AerialMPT can generalize on the other datasets, we conducted a cross-dataset validation of \NetName{} on the AerialMPT and KIT~AIS datasets. As the results in \autoref{tab:crossresults} show, the model trained on AerialMPT achieves a MOTA score of -58.9 on the test set of KIT~AIS which is 35.5 points worse than testing on AerialMPT. Nevertheless, the model trained on KIT~AIS can achieve a MOTA score of -62.8 on the test set of AerialMPT, which is 46.6 points worse than testing KIT~AIS. The results indicate that AerialMPT contains the features of KIT~AIS to a large degree, allowing the models to better generalize to various pedestrian movement scenarios. 
\CrossResults

%\begin{figure}
%\centering

%\subfigure{
%    \includegraphics[width=.4\textwidth]{figures/walkingNewSmall.png}}
%\subfigure{
%    \includegraphics[width=.4\textwidth]{figures/walkingOldSmall.png}}
    
%\caption{Illustrative tracking results on a cropped part of KIT~AIS sequence AA\_Walking\_02 with \NetName{} (top) and SMSOT-CNN (bottom). We depicted the frame numbers in the top left corner. \NetName{} is keeping trajectories significantly better.}
%\end{figure}

%\begin{figure}[t!]
%\centering
%\subfigure{
%    \includegraphics[width=.3\textwidth]{figures/karlsplatzNew.png}}

%\subfigure{
%    \includegraphics[width=.3\textwidth]{figures/karlsplatzOld.png}}
%\caption{Illustration of the better performance of SkyTrackerNet (top row) in pedestrian intersection as compared to SMSOT-CNN (bottom row). Samples are from AerialMPT (Karlsplatz sequence). The frame number is also depicted. SkyTrackerNet is able to deal with pedestrians crossing each other, even in crowded scenarios.}
%\end{figure}

%\begin{figure}
%\centering
%\subfigure[SMSOT-CNN]{
%    \includegraphics[width=.3\textwidth]{figures/wittOld.png}}

%\subfigure[SkyTrackerNet]{
%    \includegraphics[width=.3\textwidth]{figures/wittNew.png}}

%\caption{Illustrative tracking results on a cropped part of AerialMPT sequence  Witt\_3. We depicted the frame number in the top left corner. Both trackers perform poorly due to unadaptive search window size.}
%\end{figure}

%% file: sections/conclusion.tex
\section{Conclusion}\label{sec:conclusion}

In this paper, we introduced an aerial pedestrian tracking dataset, the AerialMPT dataset, and proposed \NetName, an advanced pedestrian tracking approach based on DNNs. AerialMPT is composed of 307 frames acquired from different flight campaigns over various crowd scenarios and improves the shortcomings of the only existing pedestrian tracking dataset (KIT~AIS) by better image quality and contrast, longer sequences, and a larger number of tracks. Cross-dataset validations indicate that the models trained on AerialMPT can better generalize on the other datasets.
Besides, our proposed \NetName{} is composed of an SNN, an LSTM, and a GraphCNN module to fuse appearance, temporal, and graphical features. Results on the KIT~AIS and AerialMPT dataset demonstrate that our approach successfully tackles the challenges of tracking small objects in aerial imagery, leading to a MOTA improvement by 18.2 and 13.8 compared to the baseline, respectively. Moreover, it outperforms the existing traditional and DNN-based tracking methods. Results also show that due to different GSDs and pedestrians' speeds, \NetName{} may lose pedestrians as they move out of its search window. We leave the development of an adaptive search window size for future works. Furthermore, the usage of new SNNs and different loss functions could be considered.

%% file: sections/acknowledgment.tex
\section{Acknowledgements}
We thank TernowAI\footnote{https://ternow.ai} for providing us with the high-quality labeled data.